\documentclass{article}
\usepackage{nips15submit_e,times}
\usepackage{url}
\usepackage{amssymb, amsmath, amsthm, complexity, dsfont}
\usepackage[ruled]{algorithm2e}
\usepackage{graphicx,epstopdf,subfigure,tikz,caption,float}
\usepackage{authblk}
\nipsfinalcopy

\title{Optimization for Gaussian Processes via Chaining}

\author[1]{Emile Contal}
\author[1]{C\'{e}dric Malherbe}
\author[1]{Nicolas Vayatis}
\affil[1]{CMLA, UMR CNRS 8536, ENS Cachan}

\DeclareMathOperator*{\argmin}{\arg\!\min}
\DeclareMathOperator*{\argmax}{\arg\!\max}
\def\cond{\,\vert\,}

\def\cG{\mathcal{G}}
\def\cGP{\mathcal{GP}}

\def\cN{\mathcal{N}}
\def\cO{\mathcal{O}}
\def\bR{\mathbb{R}}
\def\Var{\mathrm{Var}}
\def\cX{\mathcal{X}}
\def\one{\mathds{1}}

\def\abs#1{\left\lvert#1\right\rvert}
\def\norm#1{\left\lVert#1\right\rVert}

\def\mat#1{\mathbf{#1}}

\newtheorem{corollary}{Corollary}
\newtheorem{theorem}{Theorem}

\begin{document}

\maketitle

\begin{abstract}
In this paper, we consider the problem of stochastic optimization under a bandit feedback model.
We generalize the \textsc{GP-UCB} algorithm [Srinivas and al., 2012] to arbitrary kernels and search spaces.
To do so, we use a notion of localized chaining to control the supremum of a Gaussian process,
and provide a novel optimization scheme based on the computation of covering numbers.
The theoretical bounds we obtain on the cumulative regret are more generic
and present the same convergence rates as the \textsc{GP-UCB} algorithm.
Finally, the algorithm is shown to be empirically more efficient than its natural competitors
on simple and complex input spaces.
\end{abstract}

\section{Introduction}
% Sequential Stochastic Optimization
Optimizing an unknown and noisy function is at the center of many applications in the field of machine learning \cite{snoek2012}.
The goal of a sequential optimization procedure may be either seen
as maximizing the sum of the outputs (or rewards) received at each iteration,
that is to minimize the cumulative regret widely used in bandit problems,
or as maximizing the best reward received so far,
that is to minimize the simple regret.
This task becomes challenging when we only have mild information about the unknown function.
To tackle this challenge, a Bayesian approach has been shown to be empirically efficient
\cite{mockus1989,Jones1998,osborne2010,Hennig2012,contal2014}.
In this approach, we model the unknown function as a realization of a Gaussian Process (GP)
which allows to control the assumptions we put on the smoothness of the function
by choosing different kernels \cite{rasmussen2006}.
% Drawbacks of classical UCB
In order to prove theoretical guarantees on the regret,
algorithms in the literature typically rely on high probabilistic upper confidence bounds (UCB)
\cite{krause2011, srinivas2012, freitas2012, contal2013, djolonga2013}.
In these works and many others, the UCB is obtained with a union bound over all the points
of a discretization of the input space.
The major drawback of this approach is that the UCB depends on the cardinality of the discretization
instead of the complexity of the input space itself.
As a consequence, the convergence rates derived on the regret become arbitrary large
when the discretization becomes finer and finer.
Aiming at filling this gap between theory and practice
we propose an efficient computation of chaining for Bayesian Optimization.
Chaining, that has been recently studied in the context of sequential optimization
(see \cite{Grunewalder2010} for bandit with known horizon or \cite{Gaillard2015} in the case of online regression),
appears to be an ideal tool to capture the complexity of the search space
with respect to the smoothness of the function.
% Contribution
The contribution of this paper is twofold:
we first introduce a novel policy based on the computation of covering numbers called the {\sc Chaining-UCB} 
that can be seen as a generalization of the \textsc{GP-UCB} algorithm 
with automatic calibration of the exploration/exploitation tradeoff for arbitrary kernels and search space.
On the other hand, we provide theoretical guarantees on its regret with the same convergence rates
as its competitors, without depending on the cardinality of the discretization of the search space.
% Plan
The rest of the paper is organized as follows:
in Section~\ref{sec:algorithm}, we present the framework of our analysis,
the basic properties of GP and we introduce the \textsc{Chaining-UCB} algorithm.
In Section~\ref{sec:mainresult}, we present and discuss the upper bound on the regret.
Finally in Section~\ref{sec:expe} we compare the empirical performances
\footnote{The MATLAB code is available on \url{http://econtal.perso.math.cnrs.fr/software/}}
of the \textsc{Chaining-UCB} algorithm
to its natural competitors on simple input spaces, that is $\bR^D$,
and complex input spaces, this is directed graphs space.

\DontPrintSemicolon
\begin{minipage}[t]{.49\textwidth}
  \vspace{0pt}
  \begin{algorithm}[H]
    \caption{\textsc{Chaining-UCB}($ \cX, k(\cdot, \cdot), \eta $)}
    \label{alg:gpopt}
  %\begin{algorithmic}
    \For{$t=1,2,\dots$}{
     Compute $\mu_t$, $\sigma_t$ and $d_t$\;
     $T_0 \gets \emptyset$; $\sigma_t^\text{min} = \min_{x\in\cX}\sigma_t(x)$\;
     \For{$i=1\dots \lceil 1-\log_2(\sigma_t^\text{min}) \rceil$}{
       $\epsilon_i \gets 2^{-i+1}$\;
       $\bar{\cX} \gets \{ x\in\cX: d_t(x,T_{i-1})>\epsilon_i \}$\;
       $T_i \gets T_{i-1} \cup \textsc{Cover}(\bar{\cX}, d_t, \epsilon_i)$\;
       $H_i \gets \epsilon_i \sqrt{2\log\Big((\abs{T_i}+1) i^2 t^2 \frac{\pi^4}{\delta 6^2}\Big)}$\;
     }
     $\displaystyle x_t \gets \argmax_{x \in \cX} \mu_t(x) + \sum_{i: \sigma_t^\text{min}\leq\epsilon_i<\sigma_t(x)} H_i$\;
     Sample $x_t$ and observe $y_t$\;
   }
 %\end{algorithmic}
\end{algorithm}
\end{minipage}
\begin{minipage}[t]{.49\textwidth}
  \vspace{0pt}
  \begin{algorithm}[H]
  \caption{\textsc{Greedy-Cover}$(\cX,d,\epsilon)$}
  \label{alg:cover}
   $T \gets \emptyset$; $\bar{\cX} \gets \cX$\;
   $\forall x,x' \in \cX,~ G[x,x'] \gets \one_{d(x,x')\leq \epsilon}$\;
   \While{$\bar{\cX} \neq \emptyset$}{
     $x \gets \argmax_{x\in\bar{\cX}} \sum_{x'\in\bar{\cX}} G[x,x'] $\;
     $\displaystyle T \gets T \cup \{x\}$\;
     $\bar{\cX} \gets \bar{\cX} \setminus \{ x'\in\bar{\cX}: G[x,x'] = 1 \}$\;
   }
   \Return $T$
 \end{algorithm}
\end{minipage}

\section{The \textsc{Chaining-UCB} algorithm}
\label{sec:algorithm}

% Intro GP et GP regression
\paragraph{Bayesian optimization framework.}
Let $f:\cX \to \bR$ be the unknown function we want to optimize,
where $\cX$ is the input space which is not necessarily a subset of $\bR^D$.
We assume that $f$ is a realization of a centered Gaussian process
with known kernel $k$ satisfying $k(x,x')\leq 1$ for all $(x,x')\in\cX^2$.
To avoid measurability issues we suppose that $\cX$ is a finite set with arbitrary cardinality
(see \cite{boucheron2003, Gine2015} for more details).
A sequential optimization algorithm iterates two steps:
it first chooses $x_n$ based on $y_1, \dots, y_{n-1}$,
and next gets the noisy observation $y_n=f(x_n)+\epsilon_n$
where $(\epsilon_n)_{n\geq 1}$ are independent Gaussians $\cN(0,\eta^2)$ with known variance $\eta^2$.
Let $X_n=\{x_1, \dots, x_n\}$ be the set of queried points after $n$ iterations
and $\mat{Y}_n=[y_1, \dots, y_n]$ the associated observations packed in vector form.
Unlike the work of \cite{Grunewalder2010} this paper considers that the time horizon $n$ is unknown.
For GP, the distribution of $f$ conditioned on $\mat{Y}_n$ is a non-centered GP
of mean $\mu_{n+1}$ and kernel $k_{n+1}$ computed as follows for all $(x,x')\in\cX^2$:
\begin{equation}
  \label{eq:posterior}
  \mu_{n+1}(x) = \mat{k}_n(x)^\top \mat{C}_n^{-1}\mat{Y}_n
  \text{ ~and~ } k_{n+1}(x,x') = k(x,x') - \mat{k}_n(x)^\top \mat{C}_n^{-1} \mat{k}_n(x')\,,
\end{equation}
where $\mat{k}_n(x) = [k(x_t, x)]_{x_t \in X_n}$ is the kernel vector between $x$ and $X_n$,
and $\mat{C}_n = \mat{K}_n + \eta^2 \mat{I}$
with $\mat{K}_n=[k(x_t,x_{t'})]_{x_t,x_{t'} \in X_n}$ the kernel matrix \cite{rasmussen2006}.
We also define the variance $\sigma_n^2(x)=k_n(x,x)$ and the pseudo-distance $d_n(x,x^\star)$\,:
\begin{equation}
\label{eq:distance}
d_n(x,x^\star) = \sqrt{\sigma_n^2(x^\star) - 2k_n(x,x^\star) + \sigma_n^2(x)}\,.
\end{equation}
Note that $d_n^2(x, x^\star) = \Var[f(x^\star) - f(x) \cond X_n, \mat{Y}_{n}]$.
To measure the complexity of $\cX$ we will compute covering numbers with respect to $d_n$.

\paragraph{An upper confidence bound algorithm via chaining.}
At the core of our strategy to control the regret of the algorithm
we need an upper confidence bound (UCB) on $\sup_{x^\star\in\cX} f(x^\star)-f(x)$ for all $x\in\cX$.
A naive approach uses a union bound on $\cX$, resulting in a factor $\sqrt{\log \abs{\cX}}$
in the UCB, which is not appropriate when $\cX$ is a numerical discretization of a continuous space,
typically a multi-dimensional grid of high density.
We use the chaining trick \cite{Talagrand2014,Pollard1990,Dudley1967} to get a UCB relying
on the covering numbers of $\cX$ with respect to $d_n$ 
instead of its cardinality.
In that way the algorithm adapts to arbitrary input spaces.
The main element of our algorithm is the computation of hierarchical $\epsilon$-covers of $\cX$.
We say that a subset $T$ is an $\epsilon$-cover of $\cX$ when for all $x\in\cX$,
$d_n(x,T) \leq \epsilon$,
where $d_n(x,T)=\inf_{x'\in T} d_n(x,x')$.
The covering numbers $N(\cX,d_n,\epsilon)$ are the cardinality of the smallest $\epsilon$-cover of $\cX$
for the pseudo-distance $d_n$,
and the function \textsc{Cover}($\cX, d_n, \epsilon$) in Algorithm~\ref{alg:gpopt} returns such a set.
The \textsc{Chaining-UCB} algorithm then queries the objective function
at the point maximizing the UCB obtained by chaining.
The computation of an optimal $\epsilon$-cover is \NP-hard,
but we can easily build an efficient approximation
as shown in Algorithm~\ref{alg:cover} and discussed in Section~\ref{sec:expe}.

\section{Main results}
\label{sec:mainresult}
%\subsection{Guarantees on the regret}

\paragraph{Guarantees on the regret.}
In the following theorem we provide a high probabilistic upper bound on the instantaneous regret
incurred by Algorithm~\ref{alg:gpopt} in terms of the posterior deviations $\sigma_n(x_n)$ and covering numbers.
This inequality is used in the subsequent corollary
to obtain upper bounds on its cumulative and simple regrets.
\begin{theorem}
  \label{thm:ucb}
  For any finite $\cX$,
  let $x_1, x_2, \dots$ be the queries of the \textsc{Chaining-UCB} algorithm
  on $f$ sampled from a $\cGP(0,k)$ where $k(\cdot,\cdot)\!\leq\!1$.
  For $\delta \in (0,1)$,
  using the notations $\sigma_n=\sigma_n(x_n)$ and $c_{n,\delta}=6\sqrt{\log\frac{n^2\pi^4}{36\delta}}+15$, 
  we have with probability at least $1-\delta$ that for all $ n \in \mathbb{N}^{\star}$:
  \[\sup_{x^\star\in\cX} f(x^\star)-f(x_n) \leq
        \sigma_n \big(c_{n,\delta}-6 \log \sigma_n\big)
      + 9 \sum_{i: 2^{-i} < \sigma_n} 2^{-i} \sqrt{\log N(\cX, d_n,2^{-i})}\,.
  \]
\end{theorem}
In order to simplify this inequality and get convergence rates for the cumulative regret $R_n$ and simple regret $S_n$,
it is necessary to add some assumptions on $k$ and $\cX$.
Corollary~\ref{cor:se} gives an example of the rates we obtain
for the usual Squared-Exponential kernel $k(x,x') = e^{-\frac 1 2 \smash{\norm{x-x'}_2^2}}$
and $\cX$ in $\bR^D$.
\begin{corollary}
  \label{cor:se}
  For the SE kernel and a compact $\cX\subseteq [0,R]^D$, the \textsc{Chaining-UCB} algorithm 
  returns a sequence $x_1,x_2, \ldots$ such that
  $R_n=\cO\Big(\sqrt{n (\log n)^{D+2}}\Big)$ and
  $S_n=\cO\Big(\sqrt{\frac{(\log n)^{D+2}}{n}}\Big)$.
\end{corollary}
The proof of Theorem~\ref{thm:ucb} employs the chaining trick to get a local control on the supremum
of a non-centered GP.
Since this requires to define additional structures, the proofs are postponed to Appendix~\ref{sec:proof_thm}.
The proof of Corollary~\ref{cor:se}, in Appendix~\ref{sec:proof_cor},
first uses an upper bound on the covering numbers
via the Lipschitz property of the SE kernel.
It then applies the inequality about the mutual information proven in Theorem 5 of \cite{srinivas2012}
to obtain the given regret rates.
It is straightforward to apply this technique to other cases like linear kernels or Mat\'{e}rn kernels with parameter $\nu>2$,
since both the covering numbers and the mutual information can be bounded by similar techniques.

\begin{figure}[t]
  \centering
  \includegraphics[width=0.65\textwidth,trim={0 1cm 0 1cm},clip]{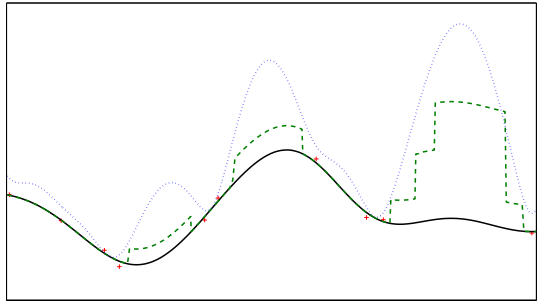}
  \caption{Illustration of the exploration/exploitation tradeoff maximized in Algorithm~\ref{alg:gpopt}.
    The red crosses are the noisy observations.
    The plain black line is the posterior mean $\mu$.
    The dashed green line is the target of the \textsc{Chaining-UCB} algorithm.
    The dotted blue line is the target used by the \textsc{GP-UCB} algorithm.
    Remark that the rectangular form is explained by the discrete sum.}
  \label{fig:ucb}
\end{figure}
\paragraph{A flexible algorithm.}
A reader familiar with classical chaining may ask why we use the bad-looking
sum $\sum_{i:\epsilon_i<\sigma_n} \epsilon_i \sqrt{\log N(\cX,d_n,\epsilon_i)}$
instead of the Dudley integral.
Even if the Dudley integral is simple to bound for certain kernel $k$ and space $\cX$,
we want Algorithm~\ref{alg:gpopt} to be able to adapt to all search space
without having to tune its parameters.
By computing the successive covering numbers, the \textsc{Chaining-UCB} algorithm
calibrates automatically the exploration-exploitation tradeoff.
This fact contrasts with previous algorithms like \textsc{GP-UCB} where the input parameter $\beta_t$
depends either on the cardinality $\abs{\cX}$ or on the Lipschitz constants
of the kernel (see Theorems 1 and 2 of \cite{srinivas2012}).
On the computational side,
the discrete sum limits the number of $\epsilon_i$-covers which need to be computed
to only the $\epsilon_i$ such that $\epsilon_i > \min_{x\in\cX} \sigma_n(x)$.
Thanks to their geometrical decay, these numbers remain low in practice.
Figure~\ref{fig:ucb} illustrates the exploration/exploitation tradeoff we obtain with this discrete sum
on a 1D toy example,
compared to the tradeoff computed with a union bound as in \textsc{GP-UCB}.
In Figure~\ref{fig:ucb} a constant term is subtracted from the UCB
in order to set the minimum of the exploration terms to zero in both \textsc{Chaining-UCB} and \textsc{GP-UCB}.

\section{Practical considerations and experiments}
\label{sec:expe}

\paragraph{Computing the $\epsilon$-covers efficiently.}
As mentioned previously the computation of an optimal $\epsilon$-cover is \NP-hard.
We demonstrate here how to build in practice a near-optimal $\epsilon$-cover using a greedy algorithm on graph.
First, remark that for any fixed $\epsilon$ we can define a graph $\cG$ where the nodes are the elements of $\cX$
and there is an edge between $x$ and $x'$ if and only if $d(x,x')\leq \epsilon$.
The size of this construction is $\cO(\abs{\cX}^2)$.
The sparse structure of the underlying graph can be exploited to get an efficient representation.
The problem of finding an optimal $\epsilon$-cover reduces to the problem
of finding a minimal dominating set on $\cG$.
We can therefore use the greedy Algorithm~\ref{alg:cover} which enjoys an approximation factor of $\log d_\mathrm{max}(\cG)$,
where $d_\mathrm{max}(\cG)$ is the maximum degree of $\cG$
(see for example \cite{Johnson1973} for a proof of \NP-hardness and approximation results).
This construction leads to an additional (almost constant) term of $\sqrt{\log \log d_\mathrm{max}(\cG)}$
in the right-hand side of Theorem~\ref{thm:ucb}.
Finally, note that this approximation is optimal unless $\P=\NP$ as shown in \cite{Raz1997}.

\begin{figure}[t]
  \centering
  \begin{subfigure}[SE kernel]{
      \includegraphics[height=.22\textwidth,trim={1.3cm .6cm 1.6cm 1cm},clip]{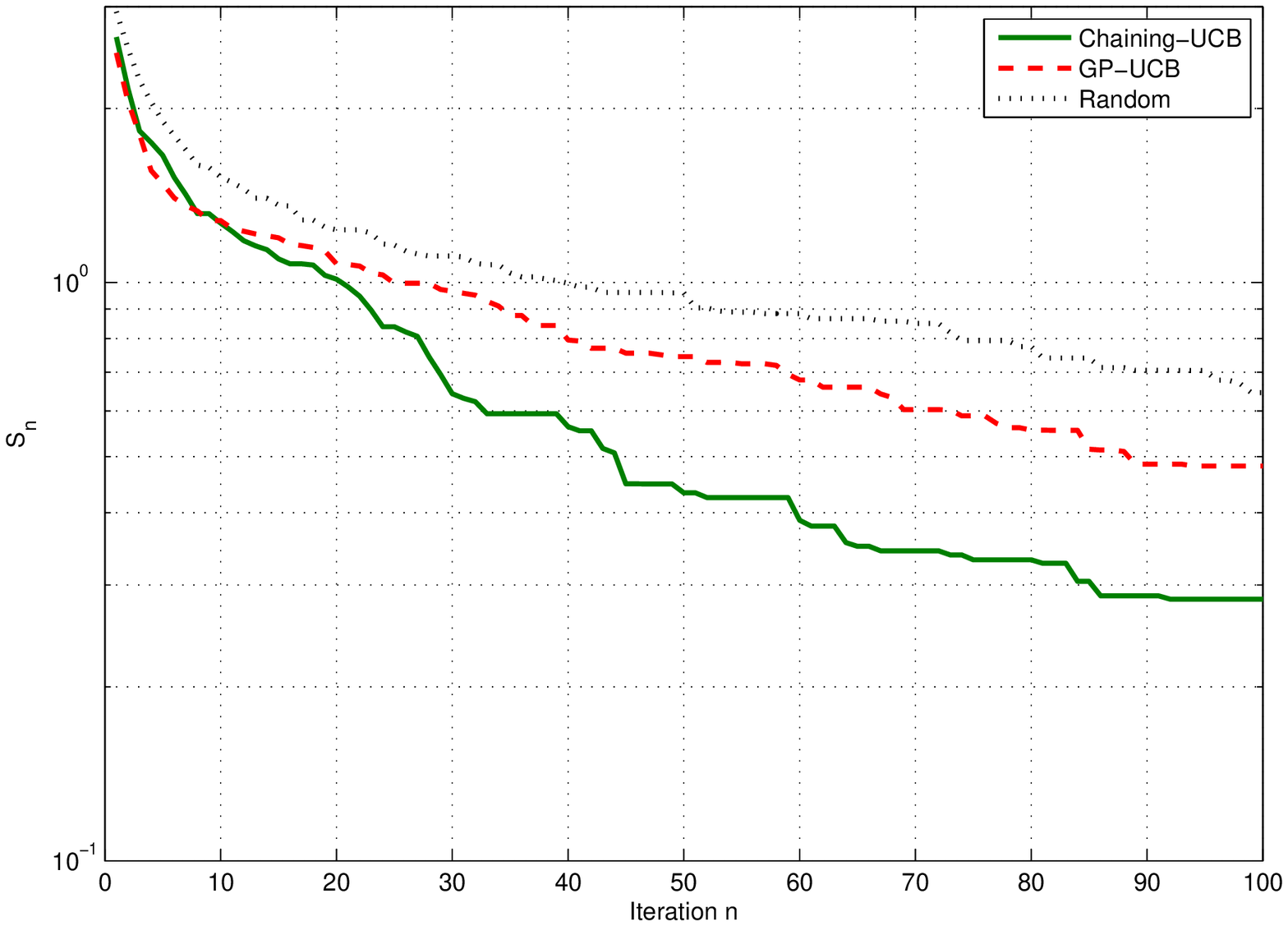}
      \label{fig:expe_d2}}
  \end{subfigure}
  \begin{subfigure}[Himmelblau]{
      \includegraphics[height=.22\textwidth,trim={1.3cm .6cm 1.6cm 1cm},clip]{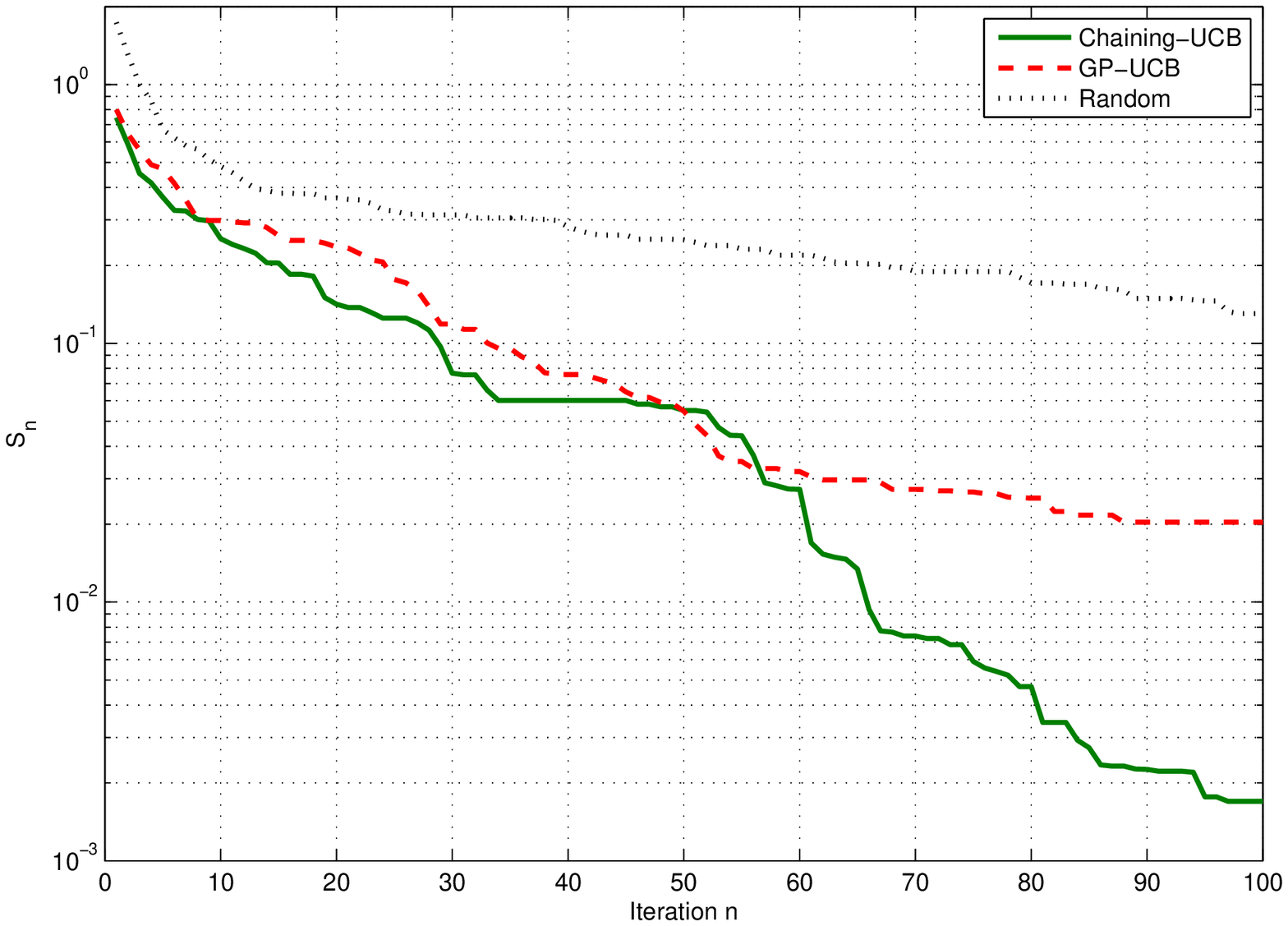}
      \label{fig:expe_him}}
  \end{subfigure}
  \begin{subfigure}[Graph kernel]{
      \includegraphics[height=.22\textwidth,trim={1.3cm .6cm 1.6cm 1cm},clip]{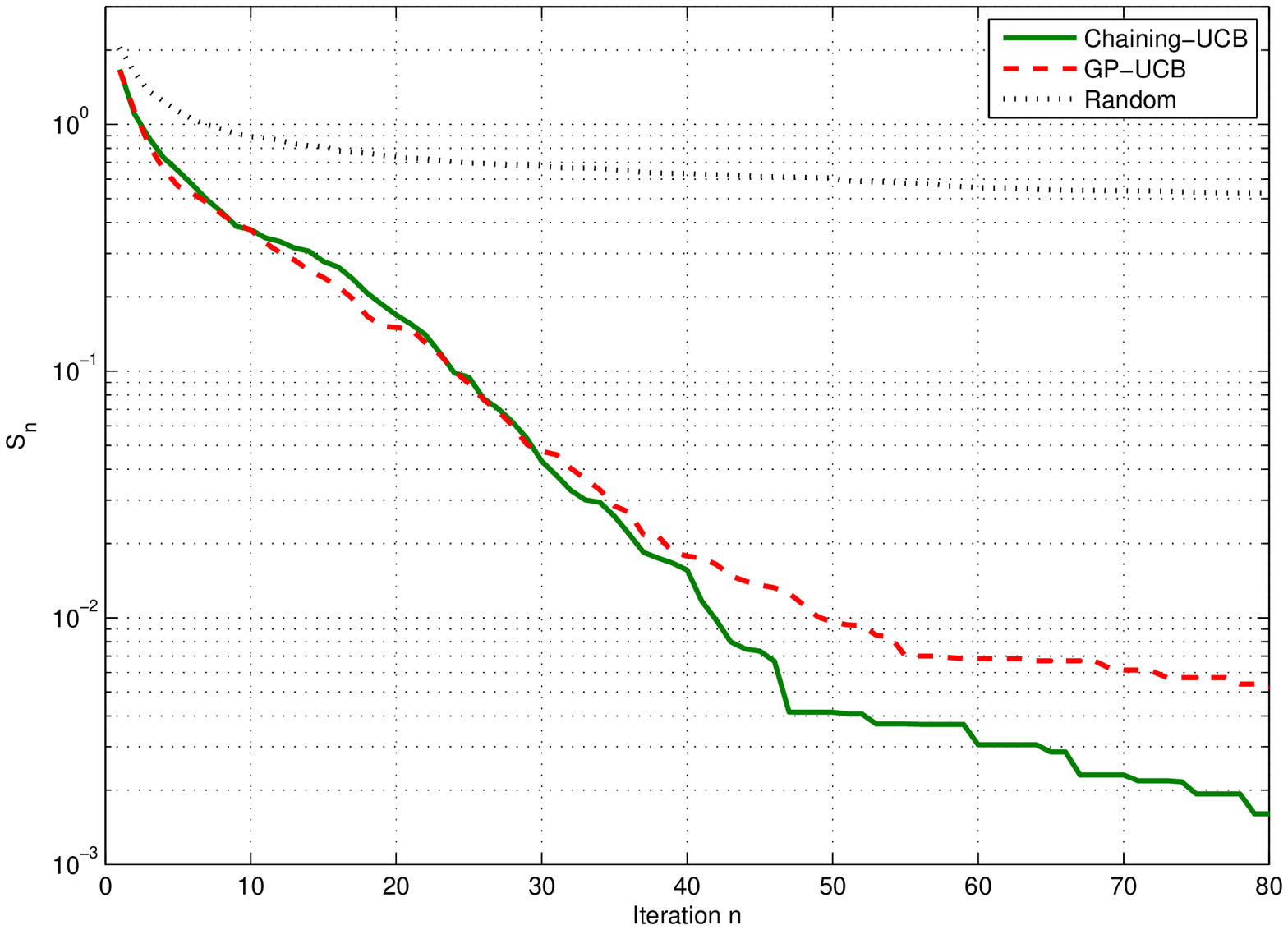}
      \label{fig:expe_graphs}}
  \end{subfigure}
  \caption{Empirical mean of the simple regret $S_n$ in terms of iteration $n$ 
    for \textsc{Chaining-UCB}, \textsc{GP-UCB} and \textsc{Random} search
  (lower is better).}
  \label{fig:expes}
\end{figure}

\paragraph{Experiments.}
In this section we compare the ability of the \textsc{Chaining-UCB} algorithm
to find the maximum of an unknown function against the \textsc{GP-UCB} algorithm from \cite{srinivas2012}
and the \textsc{Random} search.
For both the \textsc{Chaining-UCB} and the \textsc{GP-UCB} algorithms the value for $\delta$ was set to $0.05$.
The \textsc{Random} approach selects the next query uniformly among the unknown locations.
It gives a baseline to grasp the scale of the performances of both algorithms.
All three strategies are initialized with a set of $10$ noisy observations sampled uniformly over $\cX$.
Figure~\ref{fig:expes} shows the empirical mean of the simple regret $S_n$ over 32 runs.
In every experiments the standard deviation of the noise was set to $0.05$
and the search space is discrete with $\abs{\cX}=10^4$.
The SE experiment consists in generating GPs drawn from a two dimensional isotropic SE kernel,
with the kernel bandwidth set to $1$.
The search space is a uniform design in a square of length $20$.
The Himmelblau experiment is a two dimensional polynomial function
based on the Himmelblau's function with the addition of a linear trend.
It possesses several local maxima which makes the global optimization challenging.
The kernel used by the algorithm is an isotropic SE kernel
with the bandwidth chosen by maximizing the marginal likelihood of a subset of training points.
Finally we consider the task of optimizing over graphs.
Global optimization of graphs can model complex problems
as in industrial design, network analysis or computational biology.
The input space is the set of directed graphs with less than $20$ nodes.
The kernel is the shortest-path kernel \cite{Borgwardt2005} normalized such that $k(g,g)=1$ for $g\in\cX$.
Note that in this synthetic assessment we do not address the question faced in practice of choosing
the prior.
We further mention that the kernel matrix can be efficiently computed by
pre-processing all pairs of shortest paths for each graph with Floyd-Warshall's algorithm.
Figure~\ref{fig:expes} shows the \textsc{Chaining-UCB} algorithm is empirically more efficient
than the \textsc{GP-UCB} algorithm on the three test cases.
We remark that in practice, unlike \textsc{GP-UCB}, we may use a design with $\abs{\cX}\gg 10^4$
without affecting the performance of \textsc{Chaining-UCB}.
However generating a GP costs $\cO(\abs{\cX}^3)$ which limits the tractability
of synthetic assessments.

\paragraph{Conclusion.}
The theorem we derived and the experiment we performed
suggest that the automatic calibration of the exploration/exploitation tradeoff
by the hierarchical $\epsilon$-covers effectively adapts to various settings.
This chaining approach is a promising step toward generic, sound and tractable algorithms
for Bayesian optimization.

%\subsubsection*{Acknowledgments}
%
%Use unnumbered third level headings for the acknowledgments. All
%acknowledgments go at the end of the paper. Do not include 
%acknowledgments in the anonymized submission, only in the 
%final paper. 

\small
\bibliography{../../biblio/biblio}

\newpage
\appendix
\section{Proof of Theorem~\ref{thm:ucb}}
\label{sec:proof_thm}
In this section, we first demonstrate how chaining is used to analyze the supremum of a centered GP.
Since our aim is to derive an efficient algorithm, we put some effort to keep low values for the constants.
The next subsection shows how this simply extends to non-centered GP
and how to build the \textsc{UCB} algorithm with the chaining bound.
Finally we express the upper bounds in terms of the covering numbers of $\cX$.

\subsection{Using covering trees to bound the supremum of a GP}
\begin{figure}[t]
  \centering
  \begingroup
  \tikzset{every picture/.style={scale=0.8}}%
  \tikzset{
  htree leaves/.initial=2,
  sibling angle/.initial=20,
  htree level/.initial={}
}

\makeatletter

\def\htree@growth{%
  \pgftransformrotate{%
    (\pgfkeysvalueof{/tikz/sibling angle})*(-.5-.5*\tikznumberofchildren
      +\tikznumberofcurrentchild)}%
  \pgftransformxshift{\the\tikzleveldistance}%
  \pgfkeysvalueof{/tikz/htree level}%
}
\tikzstyle{htree}=[
  growth function=\htree@growth,
  sibling angle=120,
  htree level={
    \tikzleveldistance=.51\tikzleveldistance
  }
]

\makeatother

\begin{tikzpicture}[
  level distance=2cm,
  rotate=90
]
%\htree{3}
\begin{scope}[htree]
\coordinate 
child [red] {
  child [black] foreach \b in {2,3} {
    child foreach \c in {1,2,3} {
      child foreach \d in {1,2,3} {}
    }
  }
  child {
    %child [sibling angle=120, black,line width=2] {
    %  child [line width=0.5] {
    %    edge from parent node[at start,label={[label distance=2]30:$x$}] {}
    %  }
    %  child [line width=0.5] foreach \d in {2,3} {}
    %}
    child [level distance=0.9cm,sibling angle=45,black,line width=1.5,dashed,<-] { node {$x$} }
    child [sibling angle=240,black] {
      child foreach \d in {1,2,3} {}
    }
    child [sibling angle=240,line width=1.5,->] {
      child [black,line width=0.5,-] foreach \d in {2,3} {}
      child [sibling angle=260,level distance=12pt] { node [draw,shape=circle,fill=red,scale=0.2,label={[]90:$x^\star$}] {}}
    }
    child [sibling angle=240,black] {
      child foreach \d in {1,2,3} {}
    }
  }
}
child foreach \a in {2,3} {
  child foreach \b in {1,2,3} {
    child foreach \c in {1,2,3} {
      child foreach \d in {1,2,3}}}};
\node [draw,shape=circle,fill=black,scale=0.35,label={[label distance=-5]15:$x_0$}] {};
\end{scope}
\end{tikzpicture}
  \endgroup
  \caption{Illustration of the tree structure.
    The root of the tree is $x_0$, the 3 neighbors of $x_0$ form $T_1$, the next 9 neighbors of $T_1$ form $T_2$ and so on.
    The red edges connect the branch $\{p_i(x^\star)\}_{i\leq 4}$
    and the thick arrows show the points $\{\pi_i(x,x^\star)\}_{i\leq 4}$.
    Note that the positions of the nodes are arbitrary
    and do not form an $\epsilon$-cover with respect to the euclidean distance.
  }
  \label{fig:tree}
\end{figure}
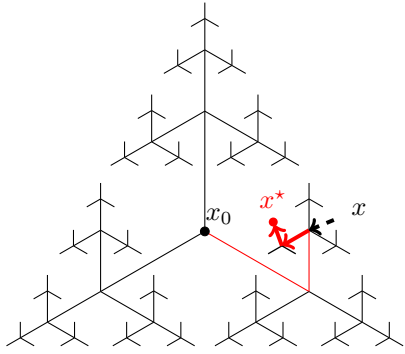

Let us fix an iteration $n$ and the observations $\mat{Y}_{n-1}$, 
we define $f_n$ the centered posterior of $f$ conditioned on $\mat{Y}_{n-1}$,
which is a Gaussian process $\cGP(0,k_n)$
where $k_n$ is defined in Equation~\ref{eq:posterior}.
By Cramer-Chernoff's method \cite{boucheron2003} we have for all $x^\star,x\in\cX$ and $u>0$ that:
\begin{equation}
\label{eq:cramer}
\Pr\big[ f_n(x^\star)-f_n(x) > d_n(x^\star,x) \sqrt{2 u} \big] < e^{-u}\,,
\end{equation}
where the pseudo-distance $d_n$ is defined Equation~\ref{eq:distance}.
We use classical chaining \cite{Talagrand2014,Dudley1967}
to obtain concentration inequalities on $\sup_{x^\star\in\cX} f_n(x^\star)-f_n(x)$
with respect to the covering numbers.
The idea of chaining performs as follows.
Let $\pi_i:\cX\times\cX \to \cX$ be a sequence of mappings such that
$\pi_0(x,x^\star) = x$ and $\pi_i(x,x^\star)\to_{i\to\infty} x^\star$ holds for all $x,x^\star\in\cX$.
We then have the telescopic sum,
\begin{equation}
\label{eq:telescopic}
\sup_{x^\star \in \cX} f_n(x^\star)-f_n(x) = \sup_{x^\star \in \cX} \sum_{i\geq 1} f_n\big(\pi_i(x,x^\star)\big)-f_n\big(\pi_{i-1}(x,x^\star)\big)\,.
\end{equation}
Since $\abs{\cX}$ may be arbitrary large we cannot perform a union bound over all $x^\star$ in Equation~\ref{eq:telescopic}.
Instead we enforce $\pi_i$ to take its values in a discrete subset $T_i$ of $\cX$
formed with carefully chosen points such that
$d(\pi_i(x,x^\star), \pi_{i-1}(x,x^\star))$ is small, so we can use Equation~\ref{eq:cramer}.
By controlling $\abs{T_i}$ we are able to achieve a union bound over $T_i$ and $i$.
We describe here the sequence $\{ T_i\}_{i\geq 0}$ of subsets of $\cX$,
which satisfy $T_0=\{x_0\}$, $T_i \subseteq T_{i+1}$ for all $i\geq 0$ and $\cX = \bigcup_{i\geq 0} T_i$.
For all level $i$ we fix an exponentially decreasing radius $\epsilon_i=2^{-i+1}$
and define $T_i$ as the smallest $\epsilon_i$-cover of $\cX$ containing $T_{i-1}$:
\begin{equation}
  \label{eq:covering_set}
  T_i \in \argmin_{\substack{T \supseteq T_{i-1}\\ d_n(x,T)\leq \epsilon_i ~\forall x\in\cX}} \abs{T}\,.
\end{equation}
The usual analysis of the chaining upper bound is not suitable for sequential optimization
since our aim is to bound $\sup_{x^\star\in\cX}f(x^\star)-f(x)$ for a specific $x$ chosen by the algorithm.
In this respect, we focus on finer results which keep track on the distance $d(x,x^\star)$.
Unfortunately, this needs to define additional theoretical quantities
and pay more attention to the sharpness of the inequalities involved.
We endow $\cX$ with a tree structure with respect to $\{T_i\}_{i\geq 0}$ and $d_n$
where the root is $x_0$ and the parent of an element of $T_{i+1}$ is its closest point among the elements of $T_i$.
Formally the parent relationship $p:\cX \to \cX$ satisfies that
for all $x \in T_{i+1}\setminus T_i,~ p(x) \in \argmin_{x' \in T_i} d(x,x')$
and $p(x_0) = x_0$ for the root.
We denote by $p_i(x)$ the ancestor of $x$ at level $i$, that is the element $x_i$ of $T_i$
such that there exists a branch $x_i,x_{i+1},\dotsc,x$ where $p(x_{j+1})=x_j$ for $j>i$.
The sequence $\{p_i(x^\star)\}_{i\geq0}$ is illustrated on Figure~\ref{fig:tree}.
The path from $x$ to $x^\star$ given by $\pi_i(x,x^\star)$ works as follows:
we start and stay on $x$ until one parent of $x^\star$ at level $i$ is not too far.
Precisely for each $x,x^\star \in\cX$ and level $i$,
$\pi_i(x,x^\star)$ is either $p_i(x^\star)$ or $x$:
\[\pi_i(x,x^\star) = \begin{cases}
p_i(x^\star) &\text{ if } \epsilon_i < d_n(x,x^\star)\,,\\
x &\text{ otherwise}\,.\end{cases}\]
Figure~\ref{fig:tree} highlights the path defined by $\pi_i(x,x^\star)$.
Using the definition of $T_i$ and $\pi_i$ we have
$d_n\big(\pi_i(x,x^\star),\pi_{i-1}(x,x^\star)\big) \leq \epsilon_{i-1}$,
and if $\epsilon_i \geq d_n(x,x^\star)$ then $\pi_i(x,x^\star)=x$ 
and the $i$-th summand in Equation~\ref{eq:telescopic} is equal to zero.
Now using the concentration inequality of Equation~\ref{eq:cramer} we get for all $u_i>0$:
\[\Pr\Big[f_n\big(\pi_i(x,x^\star)\big)-f_n\big(\pi_{i-1}(x,x^\star)\big) > \epsilon_{i-1}\sqrt{2u_i}\Big] < e^{-u_i}\,.\]
Thanks to the tree structure $\forall x\in\cX,\, i\geq 1$, $\abs{\Big\{\big(\pi_i(x,x^\star),\pi_{i-1}(x,x^\star)\big) : x^\star\in\cX \Big\}} = \abs{T_i}+1$.
We obtain for all $u_i>0$:
\[ \Pr\left[ \bigcup_{x^\star\in\cX} \bigcup_{i>0} \Big\{ f_n\big(\pi_i(x,x^\star)\big)-f_n\big(\pi_{i-1}(x,x^\star)\big) > \epsilon_{i-1} \sqrt{2u_i} \Big\}\right] < \sum_{i>0} (\lvert T_i \rvert+1) e^{-u_i}\,.\]
For any $u>0$, we set $u_i = u+\log\big((\lvert T_i \rvert+1) i^2 \frac{\pi^2}{6}\big)$,
so that $\sum_{i>0} (\lvert T_i \rvert+1) e^{-u_i} = e^{-u}$.
Combining the previous union bound with the chaining property we have a generic upper bound:
\begin{equation}
  \label{eq:chaining}
  \forall x\in\cX,\,\Pr\left[ \sup_{x^\star\in\cX} f_n(x^\star)-f_n(x) > \sup_{x^\star\in\cX} \sum_{i:\epsilon_i<d(x,x^\star)} \epsilon_{i-1} \sqrt{2 u_i}\right] < e^{-u}\,.
\end{equation}

\subsection{From classical chaining to optimization}
The previous inequality provides us with a powerful tool
to prove upper bounds on the regret.
We use two steps to be able to build the algorithm.
We first adapt this result to non-centered processes.
We then split the sum in two terms, one for $x$ and one for $x^\star$,
and we demonstrate that the queries made by Algorithm~\ref{alg:gpopt}
allow to control both terms.

At iteration $n$, the posterior process of $f$ conditioned on $\mat{Y}_{n-1}$ has a mean $\mu_n$
and covariance $k_n$ defined in Equation~\ref{eq:posterior}.
We fix $u_n>0$ and we note $\{T_i^n\}_{i\geq 1}$ the covering tree with respect to $d_n$ defined in the previous subsection.
We denote $H_i^n=\epsilon_{i-1} \sqrt{2 u_i^n}$ the associated summands of Equation~\ref{eq:chaining},
where $u_i^n=u_n+\log\big((\lvert T_i^n \rvert+1) i^2\frac{\pi^2}{6}\big)$.
In order to obtain an analogous result for $f$,
it suffices to replace $f_n(\cdot)$ in Equation~\ref{eq:telescopic} by the centered process $f(\cdot)-\mu_n(\cdot)$ 
and remark that the terms $\mu_n(\cdot)$ collide, we get:
\[
  \forall x\in\cX,\,\Pr\left[ \sup_{x^\star\in\cX} f(x^\star)-f(x)-\mu_n(x^\star)+\mu_n(x) > \sup_{x^\star\in\cX} \sum_{i:\epsilon_i < d_n(x,x^\star)} H_i^n \right] < e^{-u_n}\,.
\]
With some abuse of notation, we write in the sequel $\sigma_n=\sigma_n(x_n)$ and $\sigma_n^\star=\sigma_n(x^\star)$.
We now decompose the indices of the sum in two sets and take the supremum on both:
\[\sup_{x^\star\in\cX} \sum_{i:\epsilon_i<d_n(x,x^\star)} H_i^n 
  \leq \sup_{x^\star\in\cX} \sum_{i:\epsilon_i<\sigma_n^\star} H_i^n 
     + \sup_{x^\star\in\cX} \sum_{\substack{i:\sigma_n^\star\leq\epsilon_i\\\epsilon_i<d_n(x,x^\star)}} H_i^n\,.\]
Since for Algorithm~\ref{alg:gpopt} we have $x_n \in \argmax_{x\in\cX} \mu_n(x) + \sum_{i:\epsilon_i<\sigma_n(x)}H_i^n$, we obtain for all $x^\star\in\cX$ that
$\mu_n(x^\star)-\mu_n(x_n) \leq \sum_{i:\epsilon_i < \sigma_n} H_i^n - \sum_{i:\epsilon_i < \sigma_n^\star} H_i^n$.
We thus prove with probability at least $1-e^{-u_n}$ that:
\begin{equation}
\label{eq:two_sums}
  \sup_{x^\star\in\cX} f(x^\star)-f(x_n) \leq
      \sum_{i:\epsilon_i < \sigma_n} H_i^n 
    + \sup_{x^\star\in\cX} \sum_{\substack{i:\sigma_n^\star\leq\epsilon_i\\\epsilon_i<d_n(x_n,x^\star)}} H_i^n\,.
\end{equation}
We simplify this inequality by checking that the second sum on the right side is smaller than twice the first sum.
This involves tedious but elementary calculus using the geometric decay of $\epsilon_i$,
the linearity of $H_i^n$ in $\epsilon_{i-1}$ and the fact that $d_n(x_n,x^\star)\leq\sigma_n+\sigma_n^\star$.
It suffices to verify that for any $(a,b)\in[0,1]^2$,
$\sum_{i:a\leq 2^{-i}\leq a+b}H_i^n \leq 2\sum_{i:2^{-i}\leq b}H_i^n$. Without loss of generality we consider that $b\leq a$.
Let $i_a=-\lceil\log_2 a\rceil$, $i_b=-\lfloor\log_2 b\rfloor$ and $i_{a,b}=-\lfloor\log_2(a+b)\rfloor$.
Since $\epsilon_i=2(\epsilon_i-\epsilon_{i+1})$ and $u_i^n$ increases as $i$ increases,
we have on one side $\sum_{i=i_b}^\infty\epsilon_i \sqrt{2 u_i^n} \geq 2 \frac{b}{2} \sqrt{2 u_{i_b}^n}$,
and on the other side $\sum_{i=i_{a,b}}^{i_a}\epsilon_i \sqrt{2 u_i^n} \leq 2 b \sqrt{2 u_{i_a}^n}$.
With $u_{i_a}^n \leq u_{i_b}^n$ we get $\sum_{i=i_{a,b}}^{i_a}\epsilon_i \sqrt{2 u_i^n} \leq 2 \sum_{i=i_b}^\infty\epsilon_i \sqrt{2 u_i^n}$.
Plugging this result in Equation~\ref{eq:two_sums} we have with probability at least $1-e^{-u_n}$ that
$\sup_{x^\star\in\cX} f(x^\star)-f(x_n) \leq 3 \sum_{i:\epsilon_i < \sigma_n} H_i^n$.
If we set $u_n$ to $\log(n^2 \frac{\pi^2}{6 \delta})$ for any $\delta>0$,
that is $u_i^n = \log\left( (\lvert T_i^n\rvert+1) i^2n^2\frac{\pi^4}{\delta 6^2}\right)$,
we prove by a union bound over all $n\geq 1$ that with probability at least $1-\delta$:
\begin{equation}
\label{eq:ucb}
\forall n>0,~\sup_{x^\star\in\cX} f(x^\star)-f(x_n) \leq
3 \sum_{i:\epsilon_i < \sigma_n} \epsilon_{i-1}\sqrt{2u_i^n}\,.
\end{equation}

\subsection{Bounding with covering numbers}
\label{sec:metric_entropy}
In order to relate Equation~\ref{eq:ucb} to the size of the input space
we bound the cardinality $\abs{T_i^n}$ by the covering numbers of $\cX$.
In fact by construction of $T_i^n$ we know that
$\abs{T_i^n}+1 \leq N(\cX,d_n,\epsilon_i) + N(\cX,d_n,\epsilon_{i-1})+1$,
which is less than $3 N(\cX,d_n,\epsilon_i)$.
Note that the factor $3$ is far from being tight for large $\abs{\cX}$,
but the \textsc{Chaining-UCB} algorithm does not need this crude upper bound
since it only uses $\abs{T_i^n}$.
Splitting $u_i^n$ in three factors, the LHS of Equation~\ref{eq:ucb} is smaller than
$3 \sum_{i:\epsilon_i < \sigma_n} \epsilon_{i-1}\Big(\sqrt{2\log(\abs{T_i^n}+1)
}+2\sqrt{\log i}+\sqrt{\log \frac{n^2\pi^4}{36\delta}}\Big)$.
Using the fact that $6\sqrt{2\log 3}<9$  and that for any $s\in [0,1]$,
 $\sum_{i:2^{-i}<s} 2^{-i}\sqrt{\log i} < s-s\log s$ and $\sum_{i:2^{-i}<s} 2^{-i} < 2s$, we finally get that with probability at least $1-\delta$:
\[\sup_{x^\star\in\cX} f(x^\star)-f(x_n) \leq
9 \sum_{i:\epsilon_i < \sigma_n} \epsilon_i\sqrt{\log N(\cX,d_n,\epsilon_i)} + 6 \sigma_n \sqrt{\log\frac{n^2\pi^4}{36\delta}} + 15\sigma_n - 6\sigma_n\log \sigma_n\,,
\]
proving Theorem~\ref{thm:ucb}.

\section{Proof of Corollary~\ref{cor:se}}
\label{sec:proof_cor}
To prove Corollary~\ref{cor:se} we first remark that for the SE kernel $k(x,x')=e^{-\frac 1 2 \norm{x-x'}_2^2}$
the pseudo-distance enjoys $d(x,x') \leq \norm{x-x'}_{2}$ for all $x, x' \in \cX$.
By definition of $d_n$ we know that $d_n(x,x') \leq d(x,x')$  for all $n\geq 1$.
When $\cX\subseteq[0,R]^D$, it follows that $N(\cX, d_n, \varepsilon) \leq ( \frac{R}{\epsilon} )^D$.
Plugging this into Theorem~\ref{thm:ucb} we obtain that $R_n=\cO\Big(\sum_{t=1}^n \sigma_t \sqrt{\log t} - \sigma_t\log \sigma_t\Big)$.
We then invoke the inequality proven in Theorem 5 of \cite{srinivas2012} which shows that
$\sum_{t=1}^n \sigma_t^2 = \cO\big((\log n)^{D+1}\big)$.
We see by maximizing $R_n$ under this constraint with Lagrange multipliers
that $R_n = \cO\big(\sqrt{n (\log n)^{D+2}}\big)$.
The rate for $S_n$ directly follows from this result.

\end{document}